\documentclass[letterpaper, 10 pt, conference]{ieeeconf}  

\IEEEoverridecommandlockouts                              

\usepackage{graphics} 
\usepackage{epsfig} 
\usepackage{times} 
\usepackage{amsmath} 
\usepackage{amssymb}  
\usepackage{fixmath}
\usepackage{multirow}
\usepackage[noadjust]{cite}

\usepackage{booktabs} 
\usepackage{threeparttable}  
\usepackage{makecell}   

\usepackage{amsthm}

\let\labelindent\relax
\usepackage{enumitem}

\usepackage[linesnumbered, ruled, vlined]{algorithm2e}
\usepackage{xcolor}

\SetCommentSty{mycommfont}

\usepackage[font=footnotesize]{subcaption}
\usepackage[font=footnotesize]{caption}

\usepackage[capitalize]{cleveref}
\crefformat{equation}{(#2#1#3)}
\Crefformat{equation}{Equation~(#2#1#3)}
\Crefname{equation}{Equation}{Eqs.}

\title{\LARGE \bf
Geometric Multi-Session Map Merging with Learned Local Descriptors
}

\author{Yanlong Ma$^1$, Nakul S. Joshi$^1$, Christa S. Robison$^2$, Philip R. Osteen$^2$, Brett T. Lopez$^1$
\thanks{*This research was sponsored by the DEVCOM Army Research Laboratory (ARL) under SARA CRA
W911NF-24-2-0017. Distribution Statement A: Approved for public release; distribution is unlimited.}
\thanks{$^1$ University of California, Los Angeles, Los Angeles, CA, USA. {\tt\small \{yanlong, nakuljoshi, btlopez\}@ucla.edu}}%
\thanks{$^2$ DEVCOM Army Research Laboratory (ARL), Adelphi, MD, USA. {\tt\small \{christa.s.robison, philip.r.osteen\}.civ@army.mil}}
}

\begin{document}

\maketitle
\thispagestyle{empty}
\pagestyle{empty}
\setlength{\parskip}{0pt}
\addtolength{\topmargin}{0.05in}  

\begin{abstract}
    Multi-session map merging is crucial for extended autonomous operations in large-scale environments.
    In this paper, we present GMLD, a learning-based local descriptor framework for large-scale multi-session point cloud map merging that systematically aligns maps collected across different sessions with overlapping regions.
    The proposed framework employs a keypoint-aware encoder and a plane-based geometric transformer to extract discriminative features for loop closure detection and relative pose estimation. 
    To further improve global consistency, we include inter-session scan matching cost factors in the factor-graph optimization stage.
    We evaluate our framework on the public datasets, as well as self-collected data from diverse environments. 
    The results show accurate and robust map merging with low error, and the learned features deliver strong performance in both loop closure detection and relative pose estimation.
\end{abstract}

\section{Introduction}
\label{sec:introduction}

In recent years, the field of localization and map construction has received significant attention due to the rapid development of autonomous applications such as self-driving vehicles, infrastructure inspection, and exploration. 
An accurate and reliable map enables comprehensive environmental understanding, forming the foundation for autonomous systems to plan and navigate effectively in unknown environments.
Consequently, LiDAR-based Simultaneous Localization and Mapping (SLAM) algorithms, such as \cite{liosam,fast_lio2, dlio}, have achieved state-of-the-art performance in this field.
With consistent and reliable localization now in reach, the natural next step is to address the challenges of multi-session and/or multi-agent mapping missions aimed at building large coherent representations of complex environments with emphasis on robustness and real-time performance. 

To this end, multiple works have proposed methods for multi-session and multi-agent map merging \cite{lamp,kimera_multi,lamp2}.
In such methods, maps from multiple agents in their respective local frames are combined into one global consistent frame by identifying similar environmental information in overlapping regions, i.e., inter-session loop closures. 
This is especially challenging for large-scale mapping missions because it becomes more challenging to extract highly distinctive geometric features to accurately identify overlapping regions and estimate relative transformations across multiple maps. 
While recent approaches \cite{scan_context, overlapnet, minkloc3d} have demonstrated superior precision and robustness compared to traditional methods \cite{pfh,fpfh}, geometric feature extraction remains an open challenge due to the data sparsity (especially with LiDAR scans), variations in sensor viewpoints, high computational demands for processing dense scans, the absence of a known global reference frame, and the presence of false positives from geometrically similar features.

\begin{figure}[t!]
  \centering
  \includegraphics[width=\columnwidth]{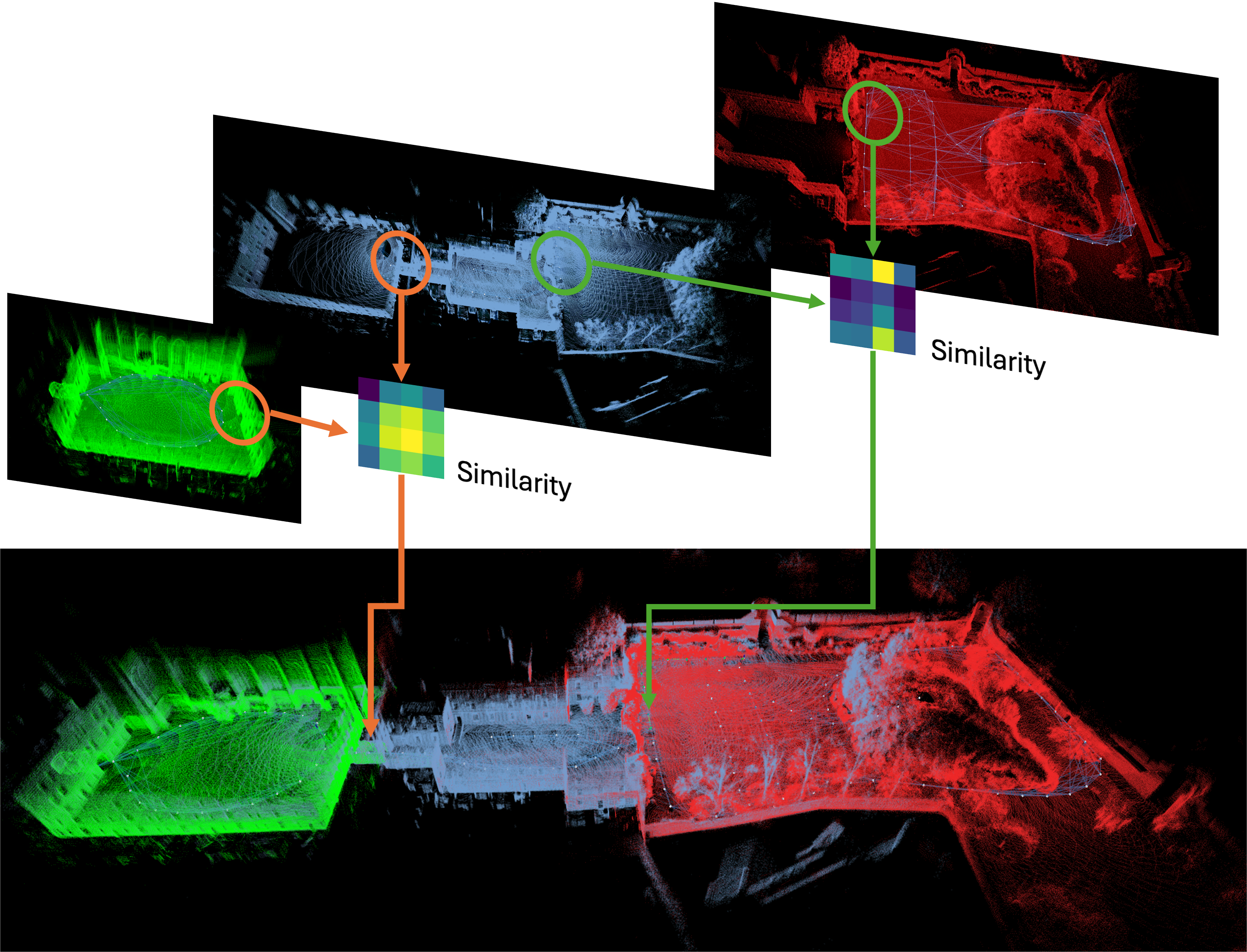}
  \caption{\textbf{GMLD} provides a robust and precise map merging framework by detecting and registering point clouds in overlapping regions through the utilization of learned discriminative descriptors and scan matching cost factors. The example shown illustrates the accurate integration of the quad and parkland with an additional corridor map from the Newer College.}
  \label{fig:cover_fig}
\end{figure}

To address this challenge, we present 3D 
\textbf{G}eometric \textbf{M}ulti-Session Map Merging with \textbf{L}earned Local \textbf{D}escriptors, a robust and precise map merging framework that generates globally-consistent maps across multiple mapping sessions or with multiple mapping agents in the same session. 
Specifically, we develop a point-based learning model capable of detecting inter-session loop closures and registering point clouds across sessions. 
Leveraging a robust outlier-rejection pipeline and a scan matching cost-aware factor-graph formulation, the proposed framework enables the online construction of large-scale maps by combining any number of mapping sessions.
The main contributions of this work are:

\begin{itemize}
\item A keypoint-aware, learning-based model that encodes the local geometry information from dense point clouds across varying viewpoints.

\item A plane-based geometric transformer encoder to strengthen the internal geometric relationships between keypoints, improving the accuracy of the merged map.

\item A factor graph formulation that supplements relative pose factors with inter-session scan matching cost minimization, enabling geometric consistency in merged maps via implicit and explicit loop closure handling.

\item A full pipeline that successfully performs 3D large-scale map merging tasks with point clouds collected by LiDAR in various environments.
\end{itemize}

We demonstrate the precision and robustness of MERRLIN by conducting a comprehensive evaluation of the proposed framework, both qualitatively and quantitatively, on the public SemanticKITTI and Newer College datasets, as well as on datasets collected at UCLA and ARL using different types of LiDAR sensors.

\section{Related Works}
\label{sec:related_works}

\subsection{Map Merging}
\label{subsec:map_merging_rel}
Map merging refers to the process of combining maps generated independently over multiple sessions or by multiple agents with no known global reference frame. 
Unlike single-agent map construction, which primarily minimizes matching cost error between consecutive point clouds, multi-session map merging aims to identify transformations between overlapping regions in different sessions and use them to align maps under a global reference frame.

Constructing a globally consistent and geometrically accurate map remains a significant challenge in large-scale environments, despite recent advances \cite{frame, automerge, lamm}.
Early studies of map merging predominantly used 2D occupancy grid maps \cite{multi_slam}, and computer vision techniques, such as the Scale-Invariant Feature Transform (SIFT), were employed to identify feature matches between sessions \cite{grid_map_sift}.
However, because occupancy grids provide only a coarse representation of the environment, the applicability of this method is inherently limited.
Recently, 3D map merging has attracted increasing attention, with many efforts focused on LiDAR-based maps directly.
SegMap \cite{segmap} applies a 3D convolutional neural network to extract features from point clouds, utilizing semantic segmentation, which is often unreliable in large-scale environments.
LAMP \cite{lamp} presents a multi-robot SLAM system that relies on ICP-based point cloud registration but lacks outlier rejection mechanisms, making merged maps error-prone. 
Kimera-multi \cite{kimera_multi} introduces a robust distributed multi-robot system with a consistent metric-semantic 3-D mesh model, leveraging depth images to extract ORB \cite{orb} features in order to carry out mesh merging. 

\subsection{Place Recognition and Registration}
\label{subsec:map_merging}
In most 3D map-merging approaches discussed above, feature extraction serves as the foundation for key components such as loop closure detection and point cloud registration between different agents.
Typically, this involves the generation of a descriptor, which is a compact representation of a point cloud or a local region within it. Some works handcraft these descriptors by utilizing analytical geometry, surface normals, curvature, or point distributions \cite{fpfh}. 
Other heuristic approaches, such as Scan Context \cite{scan_context}, construct a 2D matrix descriptor by partitioning the point cloud along azimuthal and radial directions, where each grid cell encodes height information of the contained points. 
BTC \cite{btc} proposes a combination of descriptors, including a binary descriptor encoding point occupancy along the $z$-axis and a triangle-based descriptor that represents geometric relationships among local point pairs.

Recently, learning-based approaches have dominated the study of feature extraction in the fields of both place recognition and registration. 
PointNetVLAD \cite{pointnetvlad} is an early work in point-based LiDAR place recognition, combining PointNet for local feature extraction with NetVLAD for global descriptor aggregation.
MinkLoc3D \cite{minkloc3d} applies a 3D Feature Pyramid Network to extract local features, while LoGG3D-Net \cite{logg3d} introduces a local consistency loss to enhance the local feature extraction. 
LCDNet \cite{lcdnet} designs a shared 3D voxel CNN feature extractor, which is applied in both place recognition and registration. 
Projection-based methods, like OverlapNet \cite{overlapnet} and OverlapTransformer \cite{overlaptransformer}, project the point clouds into 2D space directly, and use 2D vision techniques for feature extraction. 

\begin{figure*}[t]
  \centering
  \includegraphics[width=\textwidth]{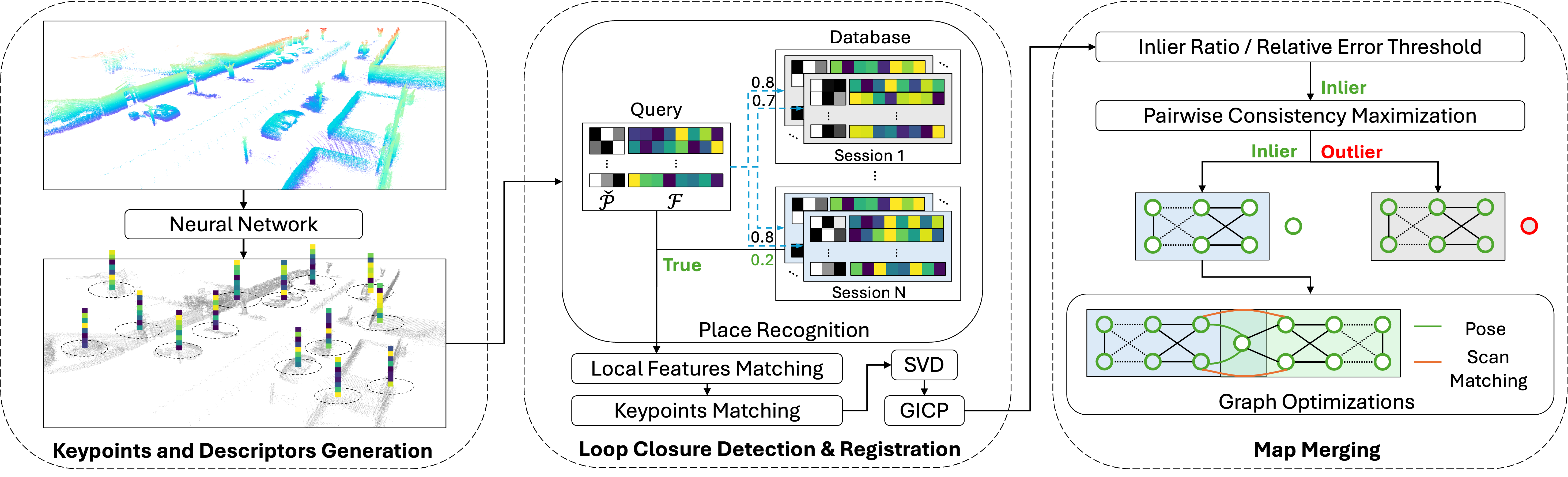}
  \caption{The proposed map merging framework consists of three modules: (i) \textit{Keypoints and Descriptors Generation} extracts keypooints and local descriptors from dense point clouds; (ii) \textit{Loop Closure Detection and Registration} identifies potential loop closures by computing the average of the smallest $n$ pairwise distances between local descriptors and estimates the relative transformation by aligning the corresponding keypoints; and (iii) \textit{Map Merging} constructs the pose and scan matching cost factor graph using candidates that satisfy the inlier ratio, relative error, and pairwise consistency maximization checks.}
  \label{fig:full_sys_arch}
\vskip -0.2in    
\end{figure*}

\subsection{Pose Graph Optimization}
\label{subsec:pose_graph_optimization}
Pose graph optimization is widely used to handle loop closures in both single- and multi-session SLAM.
In this framework, discrete poses within a session are constrained by relative pose measurements obtained from odometry and loop closure detections, typically modeled as Gaussian distributions. 
By solving for the set of poses that maximize posterior probability, trajectory estimates from one or multiple sessions can be refined to achieve global consistency within a unified coordinate frame.
Existing approaches for map merging represent inter-session loop closures as relative pose constraints between keyframes from different sessions. 
For example, \cite{lamp} and \cite{automerge} use relative pose constraints to connect subgraphs from multiple sessions, enabling the construction of a consistent global map.
However, relative pose constraints, which are typically obtained by minimizing a scan-matching cost function, are poorly represented by the unimodal Gaussian distributions that model them in factor graphs.
To address this limitation, global matching cost minimization, which employs factors to minimize scan-matching cost between overlapping LiDAR scans, has been explored as a way to ensure that the pose graph optimization promotes a geometrically consistent well-aligned map \cite{matching_cost_factor}.

\section{Methods}
\label{sec:methods}

In this section, we detail our proposed framework for merging LiDAR maps from multiple mapping sessions or from other mapping agents.
The framework consists of three components: i) a keypoint-aware downsampling strategy, ii) a plane-based geometric self-attention module that enhances the descriptiveness of local feature extraction, and iii) a scan matching cost aware inter-session pose-graph optimization that produces a globally consistent merged map.
We first present an overview of the complete pipeline, followed by detailed descriptions of each module.

\subsection{System Overview}
\label{subsec:system_overview}
An overview of the proposed pipeline is shown in \cref{fig:full_sys_arch}. 
The inputs into the proposed pipeline are keyframes---a subset of every acquired point cloud---from the current mapping session, along with the associated pose graph that encodes relative constraints between keyframe poses. 
In addition, the serialized pose graphs from auxiliary mapping sessions are also used as inputs for inter-session map merging.

In the keypoint and descriptor generation module, keyframes extracted by a keyframe-based SLAM algorithm (e.g., DLIO \cite{dlio,dliom}) in each session are voxelized to reduce computational complexity.
The preprocessed point clouds are then passed to a keypoint-aware learning module to extract the keypoints and corresponding features at multiple resolutions.
The self-attention module receives the downsampled keypoints with local features and applies a plane-based geometric embedding self-attention mechanism to exploit the spatial relations between keypoints.

In the loop closure detection and registration module, potential loop closures between keyframes from the current and auxiliary sessions are identified by thresholding the average of the smallest local-descriptor distances.
The initial relative transformation between the two candidate keyframes is then estimated by minimizing the distance between keypoints associated with descriptor correspondences and further refined using GICP \cite{gicp}. 

In the map merging module, false loop-closure candidates are subsequently rejected by verifying that both the inlier ratios and the relative pose errors—quantified by the Mahalanobis distance produced by GICP—remain within acceptable thresholds.
Outliers are further removed by enforcing pairwise consistency and selecting the maximum clique of consistent inter-session loop closures \cite{pcm}. 
Finally, the validated inter-session loop closures are inserted as relative pose factors between the factor graphs of the current and auxiliary sessions, while additional scan matching cost factors are placed between inter-session keyframe pairs in overlapping regions. 
Optimizing the resulting pose graph produces a geometrically consistent and globally aligned merged map.

\subsection{Keypoint-Aware Local Descriptor Generation}
\label{subsec: keypoint_aware_local_descriptor_generation}
\begin{figure}[t]
  \centering
  \includegraphics[width=\columnwidth]{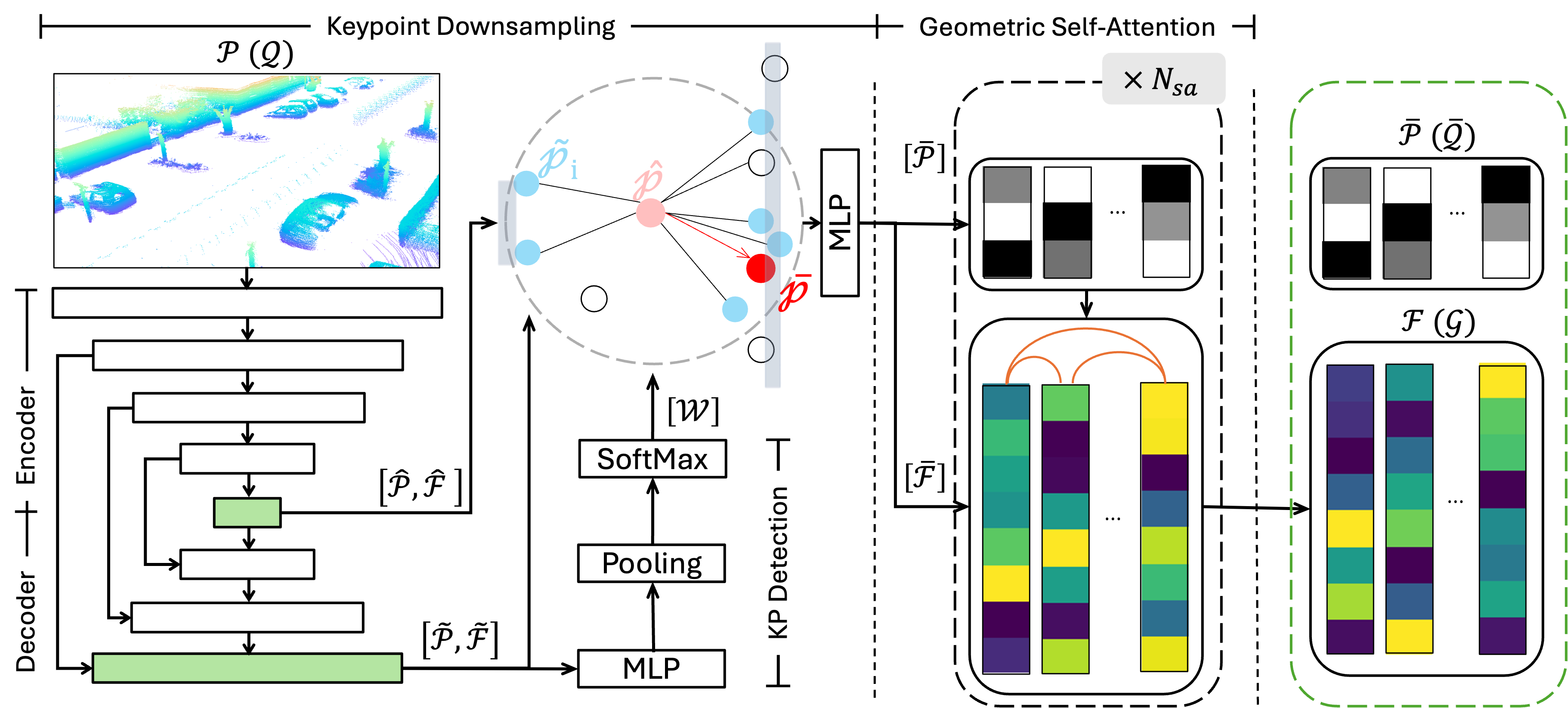}
  \caption{The neural network comprises two modules: (i) a \textit{Keypoints-Aware Downsampling} module that identifies consistent keypoints and their associated local descriptors, and (ii) a \textit{plane-based geometric self-attention} module that enhances the descriptiveness of the extracted local features.}
  \label{fig:nn_structure}
\end{figure}

The objective of generating local descriptors is to obtain a compact and discriminative representation of a local region in a point cloud. 
A locally consistent keypoint representation not only reduces variability in the neighborhood used for local descriptor generation but also decreases the Euclidean error during keypoint correspondence estimation, thereby improving the accuracy of the relative transformation estimation between keyframes.
Therefore, we propose a KPConv-based, keypoint-aware local descriptor generation framework \cite{kpconv} that incorporates a keypoint detection module to produce consistent keypoints and local descriptors.
The network structure is illustrated in \cref{fig:nn_structure}.
We utilize the backbone to extract multi-level features for the corresponding points hierarchically. 
Point clouds received from LiDAR are usually very dense and contain redundant point-wise correspondences.
To carry out real-time pairwise distances computing, point clouds must be downsampled into a sparse subset of points.
However, the voxel-grid-downsampling strategy used in KPConv introduces inconsistency in the downsampled point cloud; even a small rotation of the input point cloud can alter the distribution of downsampled points and distort the geometry. 
To combat this, we introduce a geometry-dependent downsampling strategy.

\textit{Sparse Point Downsampling and Feature Extraction.} Given a point cloud $\mathcal{P} = \{ p_1, \dots, p_N \, | \, p_i \in \mathbb{R}^3\}$, we apply the KPConv-FPN backbone with fixed voxel-grid downsampling to extract features for both sparse points and dense points, doubling the voxel size in each downsampling stage. 
In the encoder's final stage, the sparse points and their corresponding features are denoted by $\hat{\mathcal{P}} = \{\hat{p}_1, \dots, \hat{p}_n \, | \, \hat{p}_i \in \mathbb{R}^3 \}$ and $\hat{\mathcal{F}} = \{\hat{f}_1, \dots, \hat{f}_{n} \, | \, \hat{f}_i \in \mathbb{R}^{\hat{d}} \}$, respectively, with $\hat{d}$ being the sparse point feature dimensionality.
In the decoder, we compute the features of upsampled points until the first downsampled layer, since the points in $\mathcal{P}$ can fully describe the geometry of the original point cloud without additional descriptors. 
We denote the upsampled dense points by $\tilde{\mathcal{P}} = \{\tilde{p}_1, \dots, \tilde{p}_m \, | \, \tilde{p}_i \in \mathbb{R}^3 \}$ and their corresponding features by $\tilde{\mathcal{F}} = \{\tilde{f}_1, \dots, \tilde{f}_{m} \, | \, \tilde{f}_i \in \mathbb{R}^{\tilde{d}} \}$, with $\tilde{d}$ being the dense point feature dimensionality.

\textit{Consistent Keypoint Detection.} We estimate the position of a new keypoint by aggregating the neighboring dense points of the sparse points generated from the downsampling branch.
Given the sparse points $ \hat{\mathcal{P}}$, we first search the $k$ nearest neighbors ($k$NN) from the dense points $\tilde{\mathcal{P}}$, giving us $|\hat{P}|$ groups, each consisting of $k$ dense points.
For each sparse point $\hat{p}_i \in \hat{\mathcal{P}}$ with a group of neighboring dense points $\{ \tilde{p}_1^{i}, \dots, \tilde{p}_k^{i} ~ | ~ \tilde{p}_j^{i} \in \tilde{\mathcal{P}} \}$ and their corresponding features $\{ \tilde{f}_1^{i}, \dots, \tilde{f}_k^{i} ~ | ~ \tilde{f}_j^{i} \in \tilde{\mathcal{F}} \}$, we compute the positions of neighboring dense points relative to each sparse points and calculate their corresponding Euclidean distance offsets.
These relative positions and distances are then concatenated with the original point features. 
This process yields $k$ neighboring points, each augmented with an enhanced feature representation in $ \mathbb{R}^{\tilde{d} + 4}$.

The weights of the $k$ neighboring points $\mathcal{W} = \{w_1, \dots, w_k \, | \, w_i \in [0,1]\}$ in each group are predicted by passing the neighboring features through a multilayer perceptron (MLP), followed by a pooling layer, and subsequently normalized with a SoftMax function.
The keypoint is computed as a weighted sum of the neighboring points in each group, as described by the equation
$
\bar{p}_i = \sum_{j=1}^{k} {w_j \, \tilde{p}_j^{i}} .
$
In addition, the corresponding feature is updated by concatenating the previous sparse point feature $\hat{f}_i$ with the weighted sum of neighboring point features, followed by an MLP. 

Intuitively, keypoints are generated by shifting the downsampled points within the convex hull defined by their neighboring dense points.
Additionally, to mitigate overfitting, we employ random dilation grouping as in \cite{rskdd}. 
Rather than selecting the $k$ nearest neighbors of each sparse point, we first identify the $2k$ nearest neighbors and randomly sample $k$ points in each iteration. 
Completing this pipeline, we obtain the keypoints $\bar{\mathcal{P}}$ and their corresponding local descriptors $\bar{\mathcal{F}}$ as a compressed representation of the point cloud.

\subsection{Plane-Based Geometrical Transformer Encoder}
\label{subsec:plan_based_geometrical_transformer_encoder}
The local descriptors $\bar{\mathcal{F}}$, computed as previously described, encode the local geometric information around the keypoints accurately, but they do so without considering the high-level structure of the point cloud.
To combat this, we apply a plane-based geometrical transformer encoder design to enhance the local descriptors with the additional geometric information between keypoints.

Inspired by the \cite{geo-transformer}, which encodes the Euclidean distances and angles between points, we design a plane-based transformer encoder that not only considers the point-to-point but also the plane-to-plane geometric information of the keypoints.
The proposed transformer encoder is comprised of $N_{sa}$ independently configured plane-based geometric self-attention modules. 
Each module employs the conventional self-attention structure augmented with a planar geometric embedding component.
We consider two metrics to represent the planar relationship between the keypoints: Mahalanobis distance and surface angle.

\textit{Mahalanobis Distance Embedding.} Given any two keypoints $\bar{p}_i$, $\bar{p}_j \in \bar{\mathcal{P}}$, and the dense point cloud ${\tilde{\mathcal{P}}}$, we first compute the covariance matrix of the $20$ nearest dense points to the keypoints.
We denote the two covariance matrices as $\Sigma_i$ and $\Sigma_j$.
The Mahalanobis distance can be computed with the equation 
$d_{ij} = (\bar{p}_i - \bar{p}_j)^{T} (\Sigma_i + \Sigma_j)^{-1} (\bar{p}_i - \bar{p}_j)$, and the embedding $\varepsilon_{ij}^M \in \mathbb{R}^{\bar{d}}$ via the sinusoidal function
\begin{equation}
\label{eq:sinusoidal_function}
   \varepsilon_{ij, 2k}^{M} = \sin{\left(\beta_d ~ d_{ij}/c_k \right)} ~~~ \varepsilon_{ij, 2k+1}^{M} = \cos{\left(\beta_d ~ d_{ij}/c_k \right)}
\end{equation}
where the hyper-parameter $\beta_d$ controls the sensitivity of the Mahalanobis distance and $c_k = 10000^{2k / \bar{d}}$ with $\bar{d}$ being the dimension of the feature vector.

\textit{Surface Angle Embedding.} Given a covariance matrix associated with a keypoint, we consider the eigenvector with the smallest eigenvalue of the covariance matrix to be the surface normal. 
With the smallest eigenvalues of covariances $\Sigma_i$ and $\Sigma_j$ being $n_i$ and $n_j$, we compute the angle of the surface normal as
$\alpha_{ij} = \arccos{(n_i^\top n_j)}$. 
The embedding ${\varepsilon}_{ij}^{N}$ can then be computed using the same formulation as in \cref{eq:sinusoidal_function}, with the angle $\alpha_{ij}$ and hyper-parameter $\beta_n$ replacing $d_{ij}$ and $\beta_d$, respectively.

We also include the point-wise Euclidean distance embedding ${\varepsilon}_{ij}^{E}$ and triplet-wise angular embedding ${\varepsilon}_{ij}^{A}$  proposed in \cite{geo-transformer}.
Finally, the geometric information between keypoints $\bar{p}_i$ and $\bar{p}_j$ in the point cloud is represented as the combination of all embeddings
\begin{equation*}
{\varepsilon}_{ij} = {\varepsilon}_{ij}^{M} {W}^{M} + {\varepsilon}_{ij}^{E} {W}^{E} + {\varepsilon}_{ij}^{N} {W}^{N} + {\varepsilon}_{ij}^{A} {W}^{A},
\end{equation*}
where ${W}^{M}$, ${W}^{E}$, ${W}^{N}$, ${W}^{A} \in \mathbb{R}^{\bar{d} \times \bar{d}}$ are the respective projection matrices of Mahalanobis distance, Euclidean distance, surface angle, and triplet-wise angle embeddings. 
Given the keypoints $\bar{\mathcal{P}}$ and their corresponding features $\bar{\mathcal{F}}$, the new geometric-context aware local descriptor $f_i$ of the keypoint $\bar{p}_i$ can be computed using this equation
\begin{equation*}
f_i = \sum_{j=1}^{|\bar{P}|} {\frac{1}{\sqrt{\bar{d}}} \, (\bar{f}_i W^Q) \, (\bar{f}_j W^K + {\varepsilon}_{ij} W^E)^\top \, (\bar{f}_j W^V)},
\end{equation*}
where $W_Q$, $W_K$, $W_V$, and $W_E$ are the projection matrices of queries, keys, values, and embeddings.

\subsection{Loop Closure Detection and Registration}
\label{subsec:loop_closure_detection_and_registration}
Given local descriptor sets
$\mathcal{F} = \{ f_1, \dots, f_m \}$, $\mathcal{G} = \{ g_1, \dots, g_n \}$
extracted from point clouds $\mathcal{P}$ and $\mathcal{Q}$, respectively, we compute the pairwise descriptor distance matrix
$D \in \mathbb{R}^{m \times n}$, $D_{ij} = \lVert {f}_i - {g}_j \rVert_2$.
The inter-scan distance between $\mathcal{P}$ and $\mathcal{Q}$ is defined as $d(\mathcal{P}, \mathcal{Q}) = \frac{1}{s}\sum_{k=1}^{s} \mathfrak{d}_k$, where $\mathfrak{d}_k$ is the $k^{\mathrm{th}}$ smallest element of $D$ and $s$ is the number of keypoint pairs selected.
Potential loop closures between keyframes in the current session and auxiliary sessions are identified by evaluating whether the inter-scan distance between their corresponding local descriptors falls below a predefined threshold.
When multiple candidates satisfy this criterion, the keyframe with the minimum distance is selected as the loop closure candidate.

Let $\bar{\mathcal{P}} = \{ \bar{p}_1, \dots, \bar{p}_m \}$ and $\bar{\mathcal{Q}} = \{ \bar{q}_1,\dots, \bar{q}_n \}$ denote the sets of keypoints extracted from point clouds $\mathcal{P}$ and $\mathcal{Q}$, respectively. 
Following the loop closure detection scheme, initial correspondences are established by selecting $s$ keypoint pairs with the minimum descriptor distances. 
The coarse transformation $\{R, t\}$ is recovered via SVD by solving $R, t = \mathrm{arg\,min}_{R \in SO(3), t \in \mathbb{R}^3} \sum_{i=1}^{s} \| R \, \bar{p}_i + t - \bar{q}_i \|^2$.
This alignment is subsequently refined using Generalized Iterative Closest Point (GICP), initialized with the SVD solution, to achieve high-precision registration.

\subsection{Outlier Rejection}
\label{subsec:outlier_rejection}
Before merging the sequences identified by the neural network, false-positive loop closures must be eliminated, as even a single misalignment can compromise the result. 
Each loop closure candidate is evaluated by estimating the relative transformation using local descriptors and GICP. 
A candidate is rejected if the GICP alignment error exceeds a threshold or if the post-alignment inlier ratio falls below another threshold.
We further prune false positives using intra-session poses via Pairwise Consistency Maximization (PCM)\cite{pcm}, which identifies the largest subset of pairwise consistent loop closures. Consistency is verified by checking whether any two loop closures induce compatible sequence alignments under their predicted transformations. The final set of true-positive loop closures is obtained by solving a maximum-clique problem using heuristic methods.

\begin{figure}[t]
  \centering
  \includegraphics[width=\columnwidth]{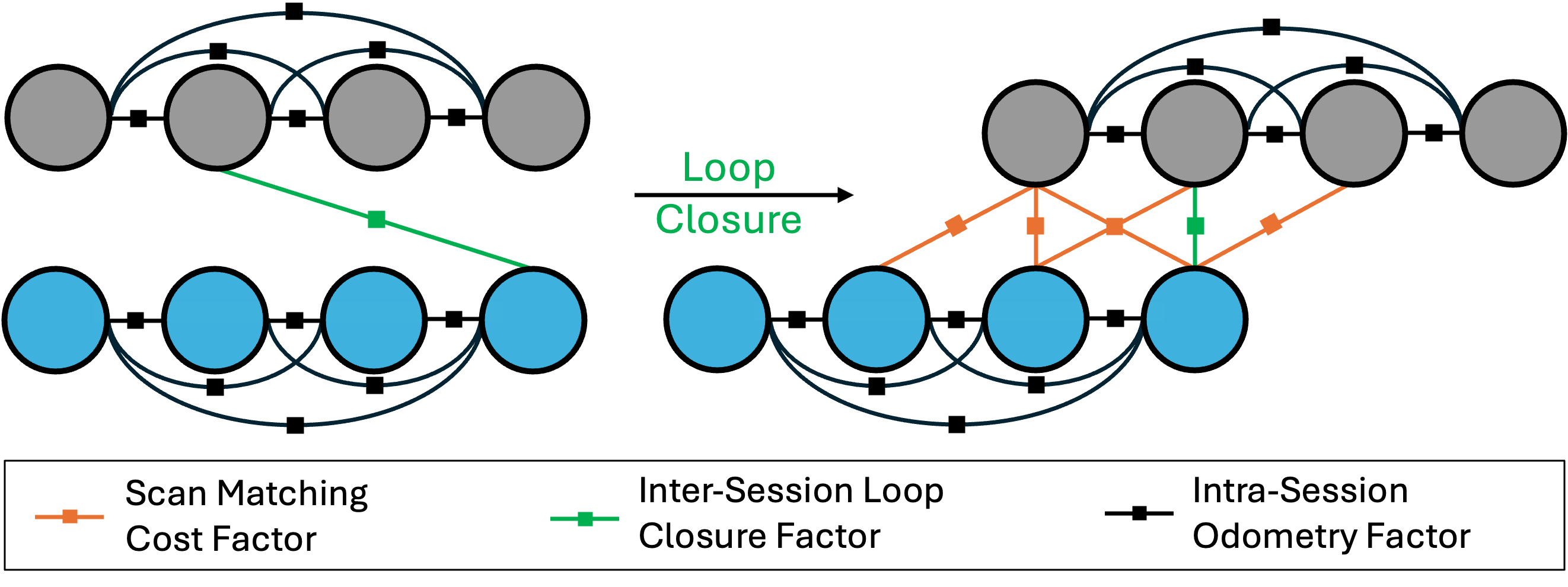}
  \caption{Each session’s pose graph, containing intra-map odometry factors (black), is serialized. When an inter-map loop closure is detected between the current and previous sessions, a loop closure factor (\textcolor{green}{green}) is added based on the relative pose. Scan matching cost factors (\textcolor{orange}{orange}) are then inserted between overlapping keyframes across sessions—both in explicit overlap regions near the loop closure and implicit overlaps resulting from the loop closure constraint.}
  \label{fig:pose_graph}
\end{figure}

\subsection{Factor Graph Optimization}
\label{factor_graph_optimization}
During each mapping session, the back-end constructs a pose graph where nodes correspond to keyframes, and between factors are added for sequentially adjacent keyframes or keyframes with sufficient overlap.
These between factors are modeled as Gaussian distributions, with covariances weighted according to the overlap between the connected keyframes \cite{dliom}.
The resulting factor graph can be optimized using a standard solver and/or serialized as an auxiliary session for inter-session merging.
Local descriptor matching provides transformations between keyframes (i.e., nodes) in the current session and those in auxiliary sessions. 

Existing approaches merge factor graphs from separate sessions by placing inter-map factors derived from relative pose estimates via place recognition \cite{automerge}. 
However, these methods cannot handle implicit loop closures, where areas mapped by multiple sessions that lack detected inter-map loop closures remain misaligned due to missing relative pose constraints.
Furthermore, representing relative pose estimates with unimodal Gaussian distributions does not adequately capture the characteristics of scan-matching solutions, which typically correspond to local minima.

To address this, we compute keyframe overlaps in separate sessions using the factor graph optimized with inter-map relative pose estimates (see \cref{fig:pose_graph}).
We place scan matching cost factors between nodes from different sessions whose keyframes overlap beyond 20\%.
We retain only the top $n_k$ keyframes with the highest overlap to maintain tractability. 

To implement these scan matching cost factors, we follow the formulation of Koide et al. \cite{matching_cost_factor}.
Let $\mathcal{P}_i$ and $\mathcal{P}_j$ be two point clouds with a set of correspondences $\mathcal{C}$, and let $\mathfrak{c} \in \mathcal{C}$.
Then, with a slight abuse of notation, we define the GICP residual as $\mathcal{E} \left( T_{ij} \mathcal{P}_i, \mathcal{P}_j \right) = \sum_{\mathfrak{c}} (p^j_\mathfrak{c} - T_{ij}\, p^i_\mathfrak{c})^\top \Omega_\mathfrak{c}^{-1} (p^j_\mathfrak{c} - T_{ij} \, p^i_\mathfrak{c})$ where $p^k_\mathfrak{c}$ is the point correspondence in cloud $k$ and $\Omega_\mathfrak{c} =  \Sigma^j_{\mathfrak{c}} + T_{ij} \, \Sigma^i_{\mathfrak{c}} \, T_{ij}^\top$ is the transformed covariance matrix with $\Sigma^i_{\mathfrak{c}}$ and $\Sigma^j_\mathfrak{c}$ are the estimated covariance matrices for the corresponding points $p_\mathfrak{c}^i$ and $p_\mathfrak{c}^j$, receptively.
Finally, $T_{ij} = T_i^{-1} T_j \in \mathbb{SE}(3)$ is the relative transformation between the point correspondences and are variable pose nodes in the factor graph to be found.  
At each linearization step in the factor graph optimization,  the residual is computed using the updated correspondences, from which a Hessian factor is derived to constrain $T_i$ and $T_j$.  
The factor is composed of Hessian matrices $H_{ii} = \sum_{\mathfrak{c}} A_\mathfrak{c}^\top \Omega_\mathfrak{c} A_\mathfrak{c}$, $H_{ij} = \sum_{\mathfrak{c}} A_\mathfrak{c}^\top \Omega_\mathfrak{c} B_\mathfrak{c}$, and $H_{jj} = \sum_{\mathfrak{c}} B_\mathfrak{c}^\top \Omega_\mathfrak{c} B_\mathfrak{c}$ and vectors $b_i = \sum_{\mathfrak{c}} A_\mathfrak{c}^\top \Omega_\mathfrak{c} d_\mathfrak{c}$ and $b_j = \sum_{\mathfrak{c}} B_\mathfrak{c}^\top \Omega_\mathfrak{c} d_\mathfrak{c}$ where $A_\mathfrak{c} = \partial d_\mathfrak{c} / \partial T_i$, $B_\mathfrak{c} = \partial d_\mathfrak{c} / \partial T_j$.
Note that each Hessian and vector can be scaled by a tunable parameter that balances the weighting of GICP residuals with between-factor residuals.
We set this scaling to 0.01. 

The inclusion of these scan matching cost factors ensures that the factor graph solver enforces both the loop closures detected by the place recognition and plane-to-plane alignment in implicitly overlapping regions, even in areas without detected loop closures. 
By leveraging the conditional dependencies between factor graph variables, a solver such as iSAM2 enables efficient, real-time inference on the map.

\begin{figure*}[t]
    \centering
    \includegraphics[width=\textwidth]{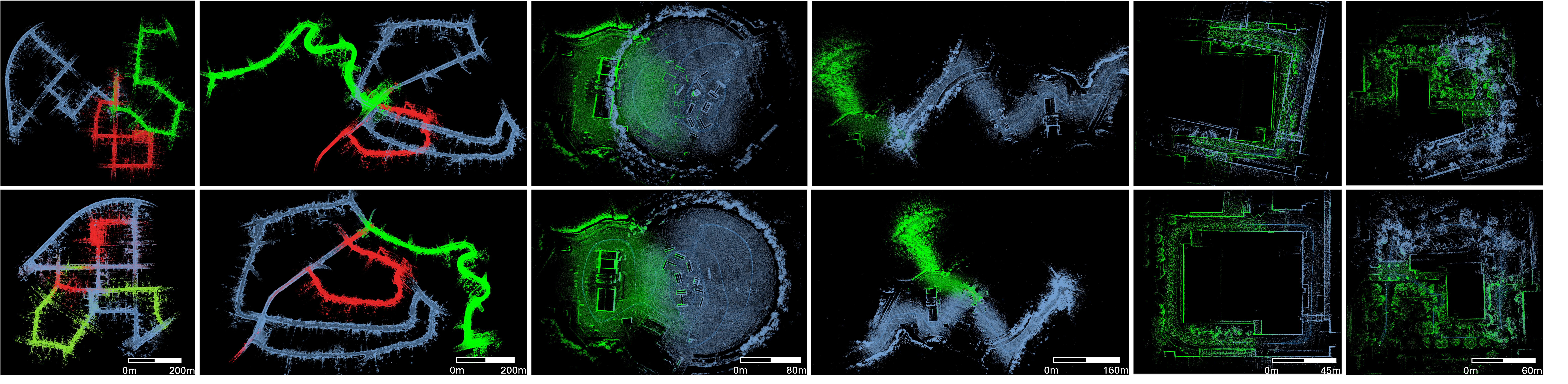}
    \caption{{Map merging results with the proposed algorithm GMLD.} The first row shows the unaligned sessions from KITTI $00$ and $02$; ARL Mout site; UCLA Engineering IV and Bunche Hall, while the second row presents the corresponding merged maps.}
    \label{fig:merged_maps}
    \vskip -0.2in
\end{figure*}

\subsection{Network Training and Loss Functions} 
\label{subsec:network_traning_and_loss_functions}
We denote the keypoints and local descriptors generated as the output of the plane-based geometric transformer as $\bar{\mathcal{P}}$, $\bar{\mathcal{Q}}$, and $\mathcal{F}$, $\mathcal{G}$, respectively. 
Correspondences between keypoints are established by matching their local descriptors. 
For descriptors in $\mathcal{P}$, the top $s$ nearest descriptors in $\mathcal{Q}$ are identified in Euclidean space. 
The rigid-body transformation between $\mathcal{P}$ and $\mathcal{Q}$ is then obtained by minimizing the distances of all descriptor correspondences using SVD.

For each keypoint, a local patch is constructed by selecting all neighboring dense points within a radius $r$. 
Ground-truth correspondences between keypoints are determined by evaluating the average geometric distance between their local patches.
Keypoints are designated as positive pairs if their corresponding local patches exhibit at least 10\% overlap; otherwise, they are treated as negative pairs.

To train the model to identify precise correspondences between local descriptors, we applied the circle loss $\mathcal{L}_{circle}$ as described in \cite{circle_loss, geo-transformer}. 
Additionally, since SVD is differentiable, we incorporated a transformation loss $\mathcal{L}_T$ that minimizes the registration error $\mathcal{L}_T = \frac{1}{|\mathcal{P}|} \sum_{p_i \in \mathcal{P}} \| (T  - \hat{T}) \, \rho_i \|^2$, where $\rho_i = (p_i,1)^\top$ and $T,\,\hat{T}\in \mathbb{SE}(3)$ are the ground-truth and predicted transformation between point clouds $\mathcal{P}$ and $\mathcal{Q}$, respectively.
We also employed the Chamfer loss $\mathcal{L}_c$ to supervise keypoint learning by minimizing the distances between corresponding keypoints.
The overall loss for local descriptor training is defined as a weighted combination of the three loss terms. Implementation details are provided in the Supplemental Materials.

\section{Experiments}
\label{sec:experiments}

\begin{table}[t]
    \centering
    \scriptsize
    \setlength{\tabcolsep}{5pt}
    \renewcommand{\arraystretch}{1.3}
    \caption{RMSE of ATE (m) on KITTI}
    \label{tab:kitti_comparison}
    \begin{tabular}{c c c c c c c}
        \hline
        \textbf{Seq} & \textbf{DLIOM} & \textbf{\makecell{FAST \\ -LIO2}} & \textbf{\makecell{DCL \\ -SLAM}} & \textbf{LAMM} & \textbf{\makecell{Ours w/o Scan \\ Matching Cost}} & \textbf{Ours} \\
        \hline
        00 & 1.498 & 3.330 & 4.182 & 2.040 & 2.185 & \textbf{1.140} \\
        02 & 5.069 & 8.749 & 18.564 & 4.746 & 6.360 & \textbf{3.709} \\ 
        05 & 1.255 & 1.765 & 2.277 & 1.639 & 1.258 & \textbf{0.964} \\
        06 & 0.772 & 0.938 & 1.690 & 0.941 & 0.833 & \textbf{0.612} \\ 
        07 & 0.668 & 0.956 & Fail & \textbf{0.561} & 0.684 & 0.713 \\ 
        \hline
    \end{tabular}
\end{table}

In this section, we comprehensively evaluate the accuracy and robustness of the proposed method both qualitatively and quantitatively, using the public KITTI and Newer College datasets and two self-collected datasets from UCLA and ARL's Graces Quarters facility. 
We compare our approach against other multi-session map merging algorithms and single-session SLAM algorithms.
Moreover, we demonstrate the effectiveness of the learned descriptors on place recognition and registration. 

\subsection{Implementation and Training Details}
\label{subsec:implementation_and_training_details}
We train the model with sequences $00$ - $09$ of KITTI using the leave-one-out cross-validation strategy and evaluate it on the sequences containing loop closures ($00, 02, 05, 06, 07$). 
In addition, we assess the map merging performance on the UCLA and ARL datasets, collected with OUSTER OS-0 and OUSTER OS-1 sensors, respectively. 
The model trained on the KITTI dataset, which uses a Velodyne HDL-64E LiDAR, is employed for all these evaluations to demonstrate the robustness and generalization of the proposed framework.

The model is trained on a desktop equipped with an NVIDIA RTX 4090 GPU.
We use the ADAM optimizer with an initial learning rate of 5e{-5}, which is decayed by a factor of $0.05$ for every 7 steps. 
The local descriptors $\mathcal{F}$ have a dimension of 256.
We set $s = 256$ when selecting corresponding keypoints and computing pairwise distances to evaluate matching scores and transformations.
All the point clouds are voxelized with a leaf size of 0.3m as the inputs.
For each keypoint, the number of nearest dense points is set to $k = 64$.
The number of self-attention layers $N_{sa}$ is set to 3. 
The embedding hyper-parameters 
$\beta_m$ and $\beta_n$ are set to 
$1/4.8$ and $1/15^{\circ}$, respectively.
To mitigate overfitting, we apply data augmentation to the input point clouds in training, including random rotations of up to $\pm 180^{\circ}$ about the yaw axis and $\pm 3^{\circ}$ about the roll and pitch axes.
Random translations of up to $\pm 2\,\text{m}$, $\pm 2\,\text{m}$, and $\pm 1\,\text{m}$ are applied along the $x$, $y$, and $z$ axes, respectively. 
Additionally, $10\%$ of the points are randomly dropped, and Gaussian noise $\mathcal{N}(0, 0.01)$ is added for point jittering.

\subsection{Map Merging}
\label{subsec:map_merging_exp}
We first evaluate the performance of the proposed map merging framework qualitatively. 
We selected two sequences from the KITTI \cite{kitti_dataset} dataset ($00$, $02$), which represent the urban environment with several overlapped regions.
Each sequence is partitioned into multiple sessions such that each session maintains overlap with at least one other session.
We also included another four maps collected from ARL, UCLA, and a map of Newer College to represent the office and campus scenarios. 
Qualitatively, our framework merged all maps successfully, as seen in \cref{fig:merged_maps} and \cref{fig:cover_fig}. 
Notably, the model operates without retraining in new environments, as the local descriptors generalize more effectively than global descriptors that are typically scenario-dependent. 
In addition to the six maps presented, we evaluated the framework in several additional sessions.
These results are provided in the Supplementary Material \ref{sec:supplementary_materials}.

To assess the quantitative accuracy of the merged map, we compare our approach against the multi-session map merging algorithm LAMM \cite{lamm}, the multi-SLAM system DCL-SLAM \cite{dcl_slam}, and maps produced directly by the standard single-session SLAM algorithms, including FAST-LIO2 \cite{fast_lio2} and the backbone SLAM module DLIOM \cite{dliom} used in our framework. 
We evaluated performance using the Root Mean Square Error (RMSE) of the Absolute Trajectory Error (ATE) on five sequences of the KITTI dataset.
Note that sequence $08$ was excluded due to its large ground-truth error.

The results are shown in \cref{tab:kitti_comparison}.
Our framework achieved the highest accuracy in four of the five sequences and ranked third for sequence $07$.
Compared with DLIOM, which runs the entire trajectory without partitioning, our method attains even better performance by effectively mitigating accumulated errors.
Furthermore, \cref{tab:kitti_comparison} confirms the effectiveness of adding scan matching cost factors in improving ATE accuracy.
Note that the accuracy metrics reported for LAMM and FAST-LIO2 were taken directly from \cite{lamm} as we were unable to reproduce their results. 

\subsection{Place Recognition}
\label{subsec:place_recognition}

\begin{table*}[t]
    \centering
    \caption{Evaluation of Place Recognition on KITTI Dataset}
    \label{tab:pr_table}
    \renewcommand{\arraystretch}{1.2}   
    \setlength{\tabcolsep}{4.5pt}       
    \begin{tabular}{l cc cc cc cc cc cc}
        \hline

        \multirow{2}{*}{Sequence} &
        \multicolumn{2}{c}{KITTI 00} &
        \multicolumn{2}{c}{KITTI 02} &
        \multicolumn{2}{c}{KITTI 05} &
        \multicolumn{2}{c}{KITTI 06} &
        \multicolumn{2}{c}{KITTI 07} &
        \multicolumn{2}{c}{Average}\\
        \cmidrule(lr){2-3} \cmidrule(lr){4-5} \cmidrule(lr){6-7} \cmidrule(lr){8-9} \cmidrule(lr){10-11} \cmidrule(lr){12-13}
        & F1 MAX & AP 
        & F1 MAX & AP
        & F1 MAX & AP
        & F1 MAX & AP
        & F1 MAX & AP
        & F1 MAX & AP \\
        \midrule

        BTC~ & 0.9670 & 0.9838 & 0.6418 & 0.5680 & 0.8895 & 0.9623 & 0.9869 & \underline{0.9926} & \underline{0.9106} & \textbf{0.9576} & 0.8792 & 0.8929 \\
        Scan Context~ & 0.9183 & 0.9388 & 0.8151 & 0.7762 & 0.8679 & 0.8699 & 0.9811 & 0.9836 & 0.4319 & 0.3861 & 0.8029 & 0.7909 \\
        OverlapTransformer~ & 0.9309 & 0.9715 & 0.8229 &0.8267 & 0.8553 & 0.9211 & 0.9544 & 0.9818 & 0.5022 & 0.4922 & 0.8131 & 0.8387 \\
        LoGG3D-Net~ & 0.9352 & 0.9733 & 0.8440 & 0.8637 & \textbf{0.9819} & \textbf{0.9918} & 0.9871 & 0.9922 & 0.8291 & 0.9038 & 0.9155 & 0.9450 \\
        LCD-Net~ & \underline{0.9693} & \underline{0.9936} & \underline{0.9186} & \underline{0.9599} & \underline{0.9503} & 0.9778 & \underline{0.9945} & \textbf{0.9939} & 0.7757 & 0.8204 & \underline{0.9217} & \underline{0.9491} \\
        Ours~ & \textbf{0.9813} & \textbf{0.9967} & \textbf{0.9518} & \textbf{0.9849} & 0.9406 & \underline{0.9788} & \textbf{1.0000} & \textbf{0.9939} & \textbf{0.9262} & \underline{0.9461} & \textbf{0.9600} & \textbf{0.9800} \\
            
        \hline
    \end{tabular}
\end{table*}

In this section, we compare the descriptiveness of the learned descriptors with several representative place recognition modules.  Specifically, we include BTC \cite{btc} used by LAMM, Scan Context used by DCL-SLAM, as well as the learning-based OverlapTransformer \cite{overlaptransformer}, LoGG3D-Net \cite{logg3d}, and LCDNet \cite{lcdnet}. 
We evaluate the performance of place recognition with KITTI sequences $00$, $02$, $05$, $06$, $07$ in terms of loop closure detection.

For each scan, we compute its similarity to all preceding scans and select the one with the minimum descriptor-space distance as the loop closure candidate.
 Precisions and recalls are computed by varying the similarity threshold.
To avoid detecting temporally adjacent scans, we excluded the most recent 50 scans from the search.
Following \cite{lamm}, we consider the detected loop closure as a true positive if the ground truth distance is less than $5$m, and as a false positive otherwise.
Additionally, as in \cite{btc}, the precision–recall curves for BTC are obtained by varying the plane overlap threshold.
For OverlapTransformer, LoGG3D-Net, and LCD-Net, we use their released pretrained models for evaluation.

\begin{figure}[t!]
    \centering
    \includegraphics[width=\columnwidth]{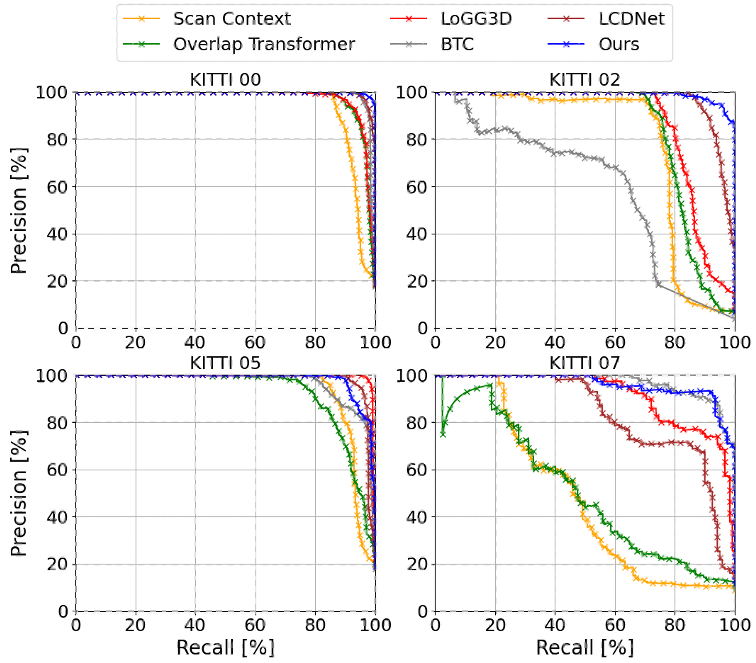}
    \caption{Precision-Recall curves for KITTI dataset. The proposed method outperforms the baselines in sequences 00, 02, and 07, while LoGG3D-Net achieves the best performance in sequence 05.}
    \label{fig:pr_curves}
\end{figure}

The precision-recall curves are shown in \cref{fig:pr_curves}, with the corresponding \textit{F1-max scores} and \textit{Average Precision (AP)} reported in \cref{tab:pr_table}.
The best method is highlighted in \textbf{bold}, and the second-best one is \underline{underlined}. 
Our method achieves the highest F1-max score across most sequences, with the exception of sequence $05$, where LoGG3D-Net attains the best performance.
Regarding Average Precision (AP), our approach ranks first on sequences $00$, $02$, and $06$, and second on the others.
Overall, our method achieves the highest average scores in both F1-max and AP, demonstrating the accuracy of the proposed descriptors.

\subsection{Relative Pose Estimation}
In this section, we evaluate the accuracy of relative pose estimation using the learned local descriptors.
Among the baseline methods, OverlapTransformer and LoGG3D-Net cannot estimate relative poses between point clouds, and Scan Context is limited to yaw estimation.
Since the released BTC code does not provide the evaluation of pose estimation, we compare our proposed method with LCDNet, which also performs both loop closure detection and pose estimation within a single network.
The RANSAC estimator in LCDNet and the GICP alignment in our method were removed to assess relative pose estimation using only local descriptors.

We evaluate pose estimation performance using all positive loop closures, specifically selecting point cloud pairs whose ground-truth distances are within 5m.
Following \cite{lcdnet}, we evaluate pose estimation using three metrics: (1) \textit{Rotation Error (RE)} -- the geodesic distance between rotation matrices, (2) \textit{Translation Error (TE)} -- the Euclidean distance of relative translation, (3) \textit{Recall} -- the ratio of successful registrations where RE $< 5^{\circ}$ and TE $<$ 2m.

The results are presented in \cref{tab:registration_table}. 
The proposed method demonstrates high accuracy in both rotation and translation estimation, achieving better results across all metrics and sequences compared with LCDNet.
In most sequences, the translation error is smaller than the voxel size (0.3m). 
We further assess performance using GICP refinement, which provides the final estimated transformation in the map merging pipeline.
With GICP refinement, the average translation error decreases from 0.247m to 0.151m, the average rotation error decreases from $0.575^\circ$ to $0.401^\circ$, and average recall improves from 99.05\% to 99.78\%.

\subsection{Ablation Studies}
\label{subsec:ablation_studies}
We assessed the contributions of the proposed framework via ablation studies. 
All models are trained and evaluated using the same settings in Section \ref{subsec:implementation_and_training_details}. 
Sequences $05$ and $07$ are chosen due to their greater difficulty in our tests. 

\begin{figure}[t!]
    \centering
    \includegraphics[width=\columnwidth]{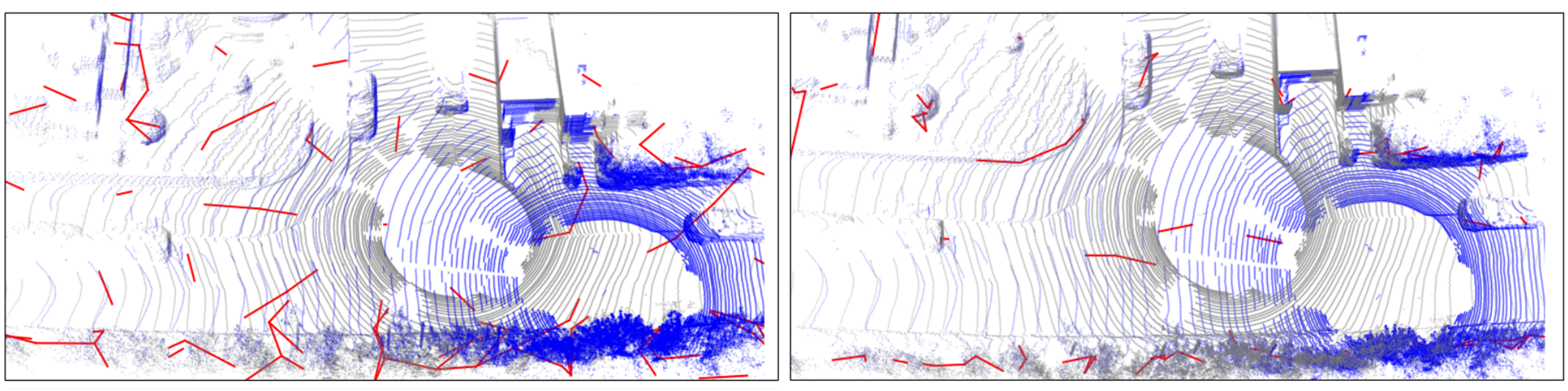}
    \caption{{Ablation study of keypoint detection.} 
    Left: No keypoint detection. Right: Matching keypoints significantly reduces the \textcolor{red}{residuals} between corresponding points.}
    \label{fig:ablation_residuals}
\end{figure}

The results are reported in \cref{tab:ablation_studies_table}.
Removing the plane-based embedding leads to a clear degradation in F1 max score, confirming its effectiveness in capturing discriminative geometric structure for loop closure detection. 
Meanwhile, eliminating the keypoint detection module results in substantially larger transformation errors, demonstrating its importance for reducing registration drift.
The residuals of matched keypoints are shown in \cref{fig:ablation_residuals}, with a clear reduction when the keypoint module is included.

\begin{figure}[t!]
    \centering
    \includegraphics[width=\columnwidth]{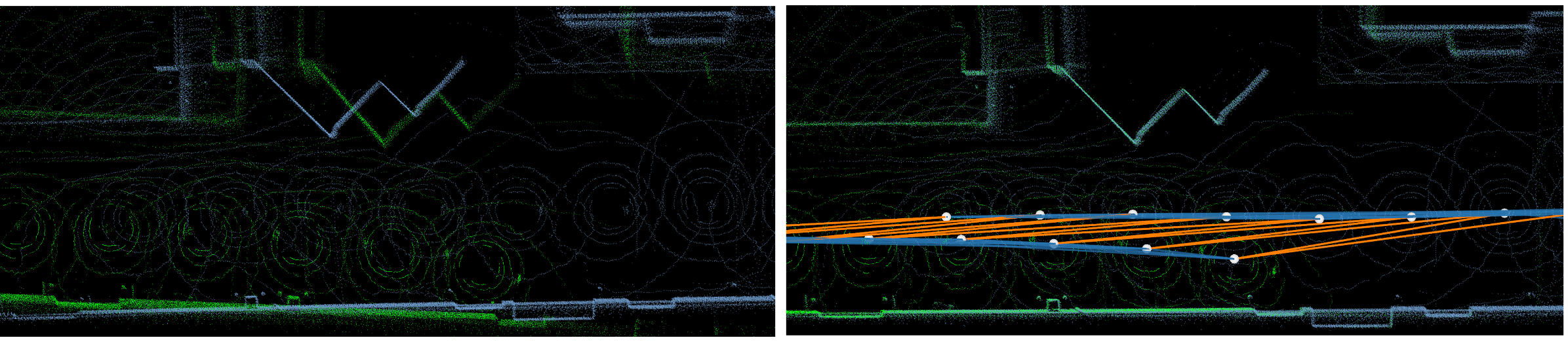}
    \caption{{Ablation study of scan matching cost factors.} Left: No scan matching cost factors. Right: Scan matching cost factors. Scan matching cost factors effectively correct local misalignments (Engineering IV Loop in \cref{fig:merged_maps}).}
    \label{fig:ablation_scan_matching}
\end{figure}

The impact of the scan matching cost factors is summarized in \cref{tab:kitti_comparison}, where all sequences exhibit notable reductions in ATE. 
As shown in \cref{fig:ablation_scan_matching}, omitting the scan matching cost factors leads to local misalignments in the merged map, further validating its effectiveness.
The orange lines represent the factor graph nodes with scan matching cost factors. 

\begin{table}[t]
    \centering
    \begin{threeparttable}
        \caption{Registration Evaluation on KITTI Sequences}
        \label{tab:registration_table}
        \setlength{\tabcolsep}{3pt}
        \renewcommand{\arraystretch}{1.2}
        \begin{tabular}{l ccc ccc}
            \toprule
            \multirow{2}{*}{Approach} &
            \multicolumn{3}{c}{LCDNet} &
            \multicolumn{3}{c}{Ours} \\
            \cmidrule(lr){2-4} \cmidrule(lr){5-7}
            & Recall\tnote{†} & TE [m] & RE\tnote{†} [deg]
            & Recall & TE [m] & RE [deg] \\
            \midrule
            KITTI 00~ & 13.35\% & 0.741 & 6.441 & 100.00\% & 0.115 & 0.391 \\
            KITTI 02~ & 4.17\% & 1.136 & 7.476 & 97.66\% & 0.544 & 1.095 \\
            KITTI 05~ & 15.02\% & 0.711 & 6.264 & 98.89\% & 0.236 & 0.655 \\
            KITTI 06~ & 10.59\% & 1.359 & 6.565 & 100.00\% & 0.091 & 0.268 \\
            KITTI 07~ & 16.02\% & 1.083 & 6.613 & 98.71\% & 0.250 & 0.469 \\
            \bottomrule
        \end{tabular}
        \begin{tablenotes}
            \item[†] The original implementation reported only the yaw error; we provide full rotation error.
        \end{tablenotes}
    \end{threeparttable}
\end{table}

\begin{table}[t]
    \centering
    \scriptsize
    \caption{Ablation Studies on KITTI Sequences}
    \label{tab:ablation_studies_table}
    \setlength{\tabcolsep}{3.5pt}
    \renewcommand{\arraystretch}{1.2}
    \begin{tabular}{l ccc ccc}
        \toprule
        \multirow{2}{*}{Methods} 
        & \multicolumn{3}{c}{KITTI 05}
        & \multicolumn{3}{c}{KITTI 07} \\
        \cmidrule(lr){2-4} \cmidrule(lr){5-7}
        & \makecell{w/o \\ Keypoints} & \makecell{w/o Plane \\ Embedding} & Ours 
        & \makecell{w/o \\ Keypoints} & \makecell{w/o Plane \\ Embedding} & Ours \\
        
        \midrule
        
        F1 Max
        & 0.9318 & 0.9247 & \textbf{0.9406}
        & 0.9091 & 0.9076 & \textbf{0.9262} \\
        
        AP 
        & 0.9769 & 0.9762 & \textbf{0.9788}
        & 0.9343 & 0.9374 & \textbf{0.9461} \\
        
        Recall
        & 98.26\% & 98.86\% & \textbf{98.89\%}
        & 88.18\% & 93.65\% & \textbf{98.71\%} \\

        TE
        & 0.393 & 0.270 & \textbf{0.236}
        & 0.755 & 0.523 & \textbf{0.250} \\
        
        RE
        & 0.795 & 0.994 & \textbf{0.655}
        & 1.718 & 0.945 & \textbf{0.469} \\
        
        \bottomrule
    \end{tabular}
\end{table}

\section{Conclusion}
\label{sec:conclusion}
In this paper, we presented a large-scale multi-session map merging framework that achieves accurate and consistent map integration across sessions. 
The proposed keypoint-aware downsampling strategy enhances the consistency of local descriptors, while the plane-based geometric transformer module improves extraction accuracy by leveraging geometrical information from downsampled points. 
In addition, the scan matching cost factors effectively reduce local inconsistencies in overlapping regions, ensuring precise global map alignment. 
Experimental results validate the accuracy and robustness of the proposed framework across diverse environments. 
Notably, features extracted from point clouds are used not only for map merging, but also for a variety of other tasks in recent studies \cite{optmap}.
In future work, we plan to enhance the efficiency of the multi-stage downsampling process to reduce the computational cost of feature extraction and enable deployment in a broader range of scenarios.


\bibliographystyle{IEEEtran}
\bibliography{ref}

\section{Supplementary Materials}
\label{sec:supplementary_materials}

\subsection{Network Training and Loss Functions}
\label{subsec:network_training_and_loss_functions}
As explained in the main paper, we apply the overlap-aware circle loss outlined in \cite{circle_loss, geo-transformer}.
Given the local patches $\mathcal{O}^{P}$ of point cloud $\mathcal{P}$, we partition the patches of point cloud $\mathcal{Q}$ into two sets $\mathcal{O}_{p}^{Q}$ and $\mathcal{O}_{n}^{Q}$, where $\mathcal{O}_{p}^{Q}$ corresponds to the positive local patches and $\mathcal{O}_{n}^{Q}$ corresponds to the negative patches.
The loss of descriptors in $\mathcal{F}$ is
\begin{equation*}
\begin{split}
    \mathcal{L}_{m}^{P}
    &= \frac{1}{|O^{P}|}\sum_{o_i^{P}\in O^{P}}
    \log\Biggl[1 + \sum_{o_j^{Q}\in O_{p}^{Q}}
    e^{\gamma\,(d_i^{j}-\Delta_{p})^{2}\, \left({\lambda_i^{j}} \right)^{\frac{1}{2}}} \cdot \\
    &\qquad\qquad\qquad\qquad\qquad
    \sum_{o_k^{Q}\in O_{n}^{Q}}
    e^{\gamma\,(\Delta_{n}-d_i^{k})^{2}}\Biggr],
\end{split}
\end{equation*}
where $d_i^j$, $d_i^k$ represent the Euclidean distances between local descriptors. 
We set $\Delta_p = 0.1$, $\Delta_n = 1.4$, which are the margins of positive and negative descriptors, $\lambda_i^j$ is the overlap ratio between positive patches, and $\gamma$ serves as the hyper-parameter. 
The resulting loss of descriptors between $\mathcal{P}$ and $\mathcal{Q}$ is computed by taking the average of $\mathcal{L}_{m}^{P}$ and $\mathcal{L}_{m}^{Q} $ as 
$ \mathcal{L}_{circle} = \frac{1}{2} (\mathcal{L}_{m}^{P} + \mathcal{L}_{m}^{Q})$.

The Chamfer loss between point clouds $\mathcal{P}$ and $\mathcal{Q}$ measures the bidirectional distance between each keypoint in the downsampled sets $\bar{\mathcal{P}}$ and $\bar{\mathcal{Q}}$:
\begin{equation*}
\begin{split}
    \mathcal{L}_c (\mathcal{P}, \mathcal{Q}) 
    =& 
    \frac{1}{2|\bar{\mathcal{P}}|}
    \sum_{\bar{p}_i \in \bar{\mathcal{P}}} \min_{\bar{q}_j \in \bar{\mathcal{Q}}} || \bar{p}_i - \bar{q}_j ||_2^{\,2} \\ 
    & + \frac{1}{2|\bar{\mathcal{Q}}|}
    \sum_{\bar{q}_i \in \bar{\mathcal{Q}}} \min_{\bar{p}_j \in \bar{\mathcal{P}}} || \bar{q}_i - \bar{p}_j ||_2^{\,2}.
\end{split}
\end{equation*}
This loss encourages each keypoint in one set to be close to its nearest neighbor in the other set, providing supervision for precise keypoint learning and robust point cloud alignment.
Combined with the transformation loss introduced in the main paper, the overall training loss is defined:
\begin{equation*}
    \mathcal{L} = \beta_{circle} \mathcal{L}_{circle} + \beta_t \mathcal{L}_t + \beta_c \mathcal{L}_c,
\end{equation*}
where the weighting coefficients $\beta_{circle}, \, \beta_{t}, \, \beta_{c}$ are all set to 1.0 in the work.

\subsection{Place Recognition}
\label{place_recognition}
The proposed local descriptors are used not only to compute scan similarities but also to estimate local transformations. 
Incorporating a GICP refinement step allows us to obtain the inlier ratio between any two point clouds, which serves as an outlier-rejection criterion in our pipeline. 
Consequently, instead of generating Precision–Recall curves by varying the average descriptor distance, we can alternatively vary the inlier ratio to produce these curves.
Similar to \cite{lamm}, we select the best candidate based on the descriptor distance and use it to compute the inlier ratio for every scan.

\begin{figure}[t!]
    \centering
    \includegraphics[width=\columnwidth]{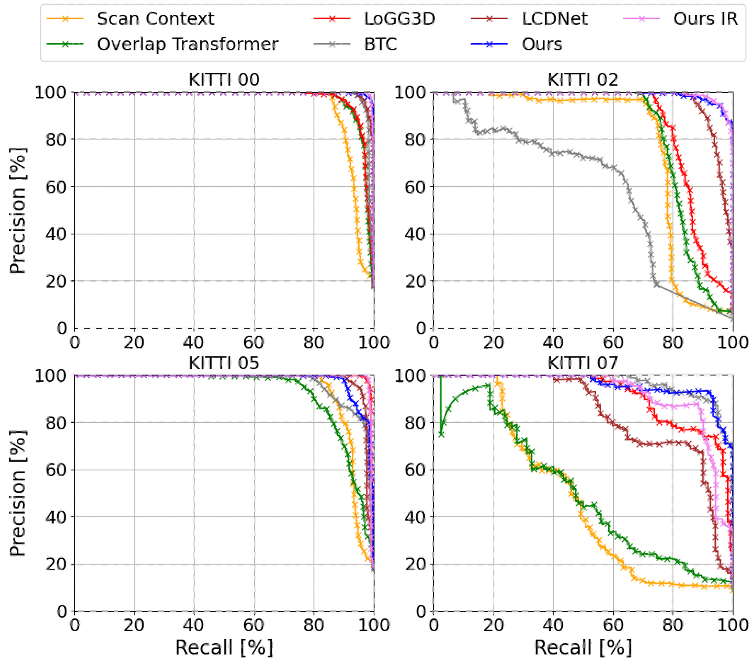}
    \caption{Precision-Recall curves of KITTI dataset (Inlier Ratio). By varying the inlier-ratio threshold, performance improves on sequences 00, 02, and 05, but degrades on sequence 07.}
    \label{fig:pr_curves_ir}
\end{figure}

The results are shown in \cref{fig:pr_curves_ir}.
The precision–recall curves show improved performance on sequences $00$, $02$, and $05$, but a degradation on sequence $07$. 
This highlights the necessity of incorporating the additional outlier-checking step based on the inlier ratio within the pipeline.
 
\subsection{Map Merging}
\label{map_merging}
\begin{figure}[t]
    \centering
    \includegraphics[width=0.96\columnwidth]{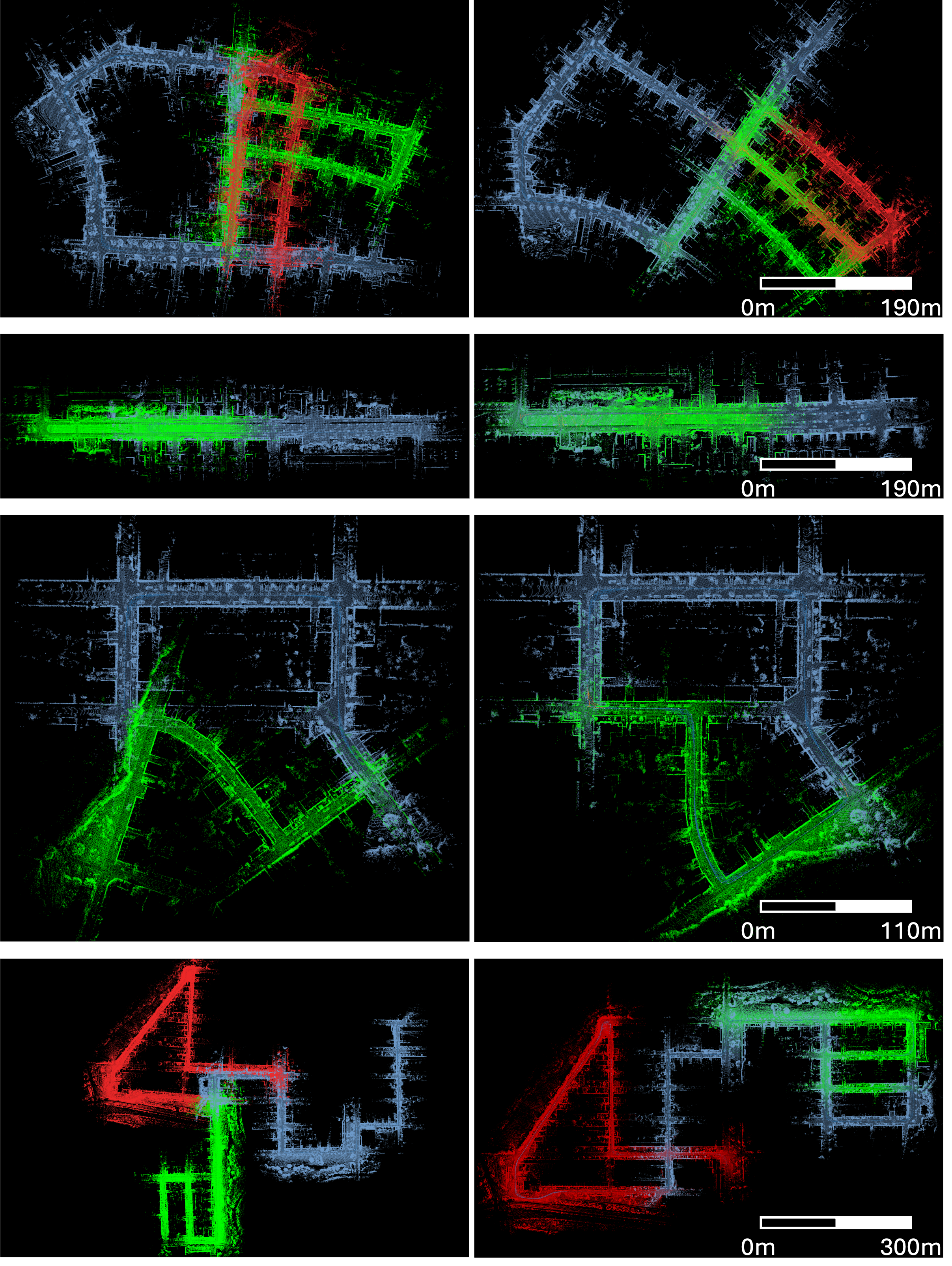}
    \caption{{Additional KITTI Map merging results}. The first column shows the unaligned sessions from KITTI 05, 06, 07, and 08, while the second column presents the corresponding merged maps.}
    \label{fig:kitti_merged_maps}
\end{figure}

\begin{figure}[t]
    \centering
    \includegraphics[width=0.96\columnwidth]{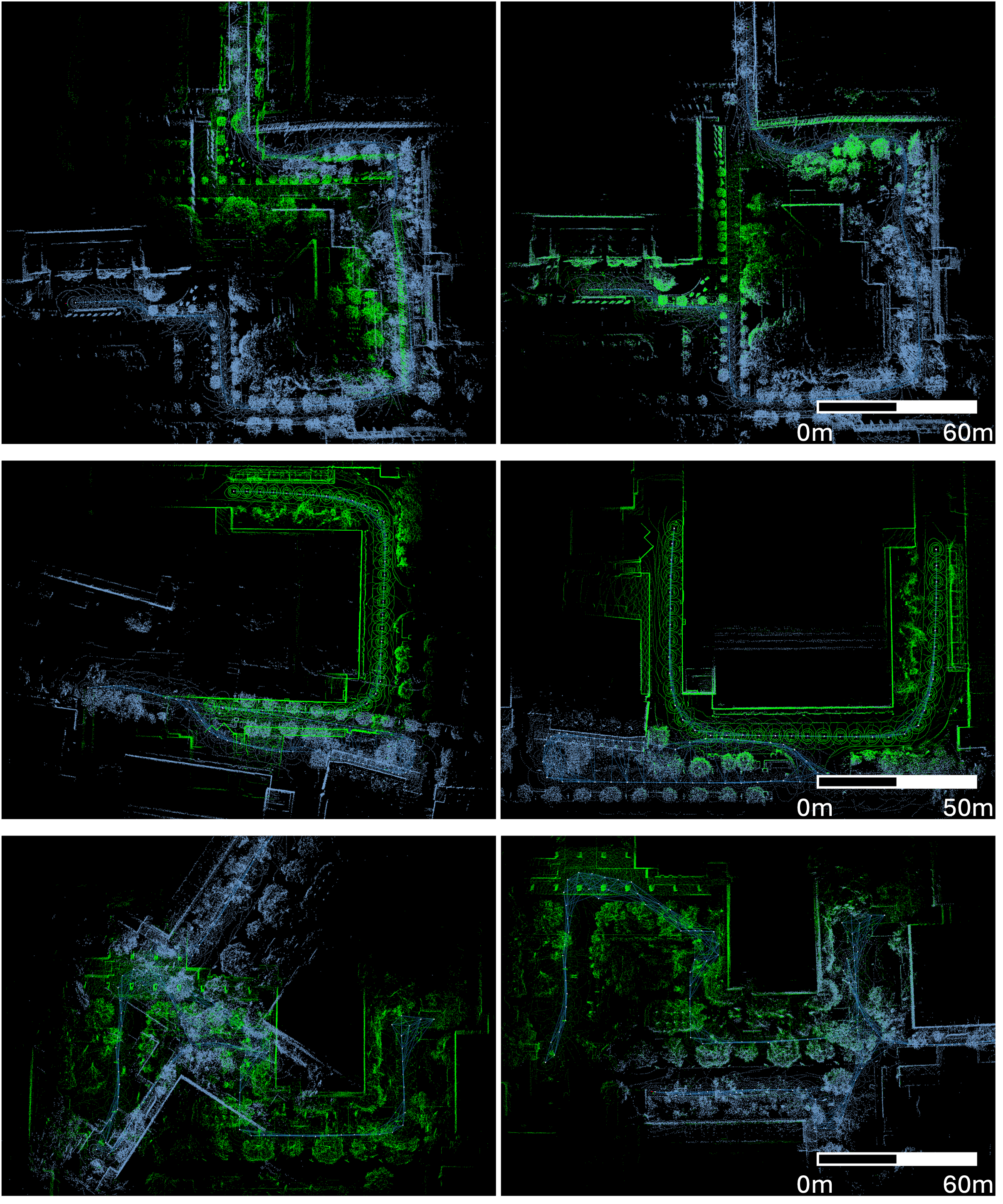}
    \caption{{Additional UCLA Map merging results}. The first column shows the unaligned sessions from UCLA West Campus, Engineering IV West, and Bunche Hall East, while the second column presents the corresponding merged maps.}
    \label{fig:ucla_merged_maps}
\end{figure}

Additional merged maps results are shown in \cref{fig:kitti_merged_maps,fig:ucla_merged_maps}.

\end{document}